%% file: main.tex
\documentclass[runningheads]{llncs}

\usepackage{eccv}

\usepackage{eccvabbrv}

\usepackage{graphicx}
\usepackage{booktabs}

\usepackage[accsupp]{axessibility}  

\usepackage[pagebackref,breaklinks,colorlinks,citecolor=eccvblue]{hyperref}
\usepackage{hyperref}

\usepackage{orcidlink}

\begin{document}

\title{Improving Virtual Try-On with Garment-focused\\ Diffusion Models} 

\titlerunning{Improving Virtual Try-On with Garment-focused Diffusion Models}

\author{Siqi Wan\inst{1 \ast} \and
Yehao Li\inst{2} \and
Jingwen Chen\inst{2}\orcidlink{0000-0002-7917-6003} \and
Yingwei Pan\inst{2}\orcidlink{0000-0002-4344-8898} \and\\
Ting Yao\inst{2} \and
Yang Cao\inst{1} \and
Tao Mei\inst{2}\orcidlink{0000-0002-5990-7307}}

\authorrunning{Siqi Wan et al.}

\institute{University of Science and Technology of China \and
HiDream.ai Inc. \\
\email{wansiqi4789@mail.ustc.edu.cn,~\{liyehao,~chenjingwen,~pandy,~tiyao\}@hidream.ai \\forrest@ustc.edu.cn,~tmei@hidream.ai}}

\maketitle

\let\thefootnote\relax\footnotetext{$^{\ast}$ This work was performed at HiDream.ai.}

\input{sec/0_abs}
\input{sec/1_intro}
\input{sec/2_relate_work}
\input{sec/3_method}
\input{sec/4_exp}

\input{sec/5_conclusion}


\clearpage  

%
%
\bibliographystyle{splncs04}
\bibliography{main}
\end{document}

%% file: sec/0_abs.tex
\begin{abstract}
  Diffusion models have led to the revolutionizing of generative modeling in numerous image synthesis tasks. Nevertheless, it is not trivial to directly apply diffusion models for synthesizing an image of a target person wearing a given in-shop garment, i.e., image-based virtual try-on (VTON) task. The difficulty originates from the aspect that the diffusion process should not only produce holistically high-fidelity photorealistic image of the target person, but also locally preserve every appearance and texture detail of the given garment. To address this, we shape a new Diffusion model, namely GarDiff, which triggers the garment-focused diffusion process with amplified guidance of both basic visual appearance and detailed textures (i.e., high-frequency details) derived from the given garment. GarDiff first remoulds a pre-trained latent diffusion model with additional appearance priors derived from the CLIP and VAE encodings of the reference garment. Meanwhile, a novel garment-focused adapter is integrated into the UNet of diffusion model, pursuing local fine-grained alignment with the visual appearance of reference garment and human pose. 
  We specifically design an appearance loss over the synthesized garment to enhance the crucial, high-frequency details. Extensive experiments on VITON-HD and DressCode datasets demonstrate the superiority of our GarDiff when compared to state-of-the-art VTON approaches. Code is publicly available at: \href{https://github.com/siqi0905/GarDiff/tree/master}{https://github.com/siqi0905/GarDiff/tree/master}.
  \keywords{Virtual Try-on \and Diffusion Model \and Appearance Prior}
\end{abstract}

%% file: sec/1_intro.tex
\section{Introduction}
\label{sec:intro}

Image-based Virtual Try-ON (VTON), a prominent research topic in computer vision field, aims to synthesize an image of a specific person wearing a desired in-shop garment. Such automatic generation of person images sidesteps the requirement of physical fitting and thus has ushered in a new era of creativity for e-commerce and metaverse. Practical VTON systems have a tremendous potential impact on real-world applications, e.g., online shopping, fashion catalog creation, etc. The objective of VTON task is three-fold: 1) human body alignment: the synthesized person image should conform to the human body/pose of given specific person; 2) garment fidelity: the synthesized person image should preserve every appearance and texture detail of the in-shop garment; 3) quality: the synthesized person image should be of high-quality with few artifacts.
\begin{center}
    \centering

    \includegraphics[width=0.95\textwidth]{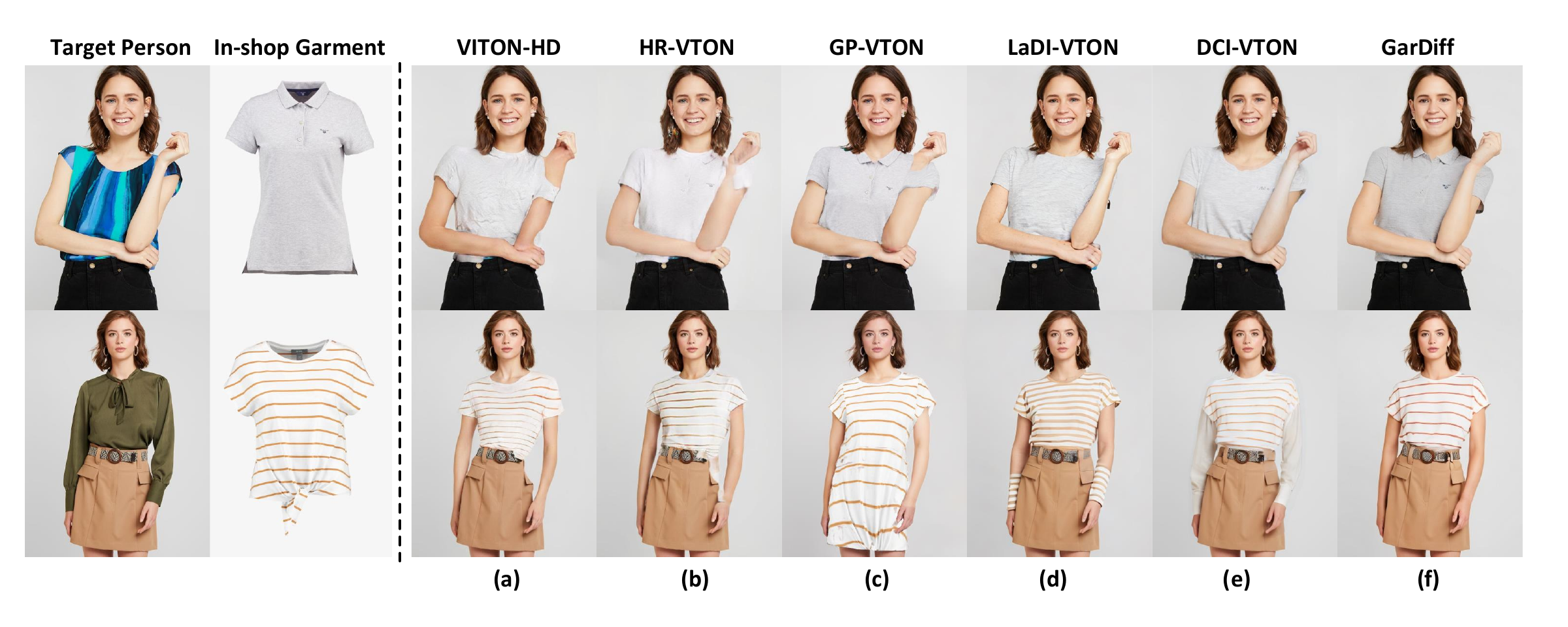}

    \captionof{figure}{Existing GAN-based VTON methods (e.g., VITON-HD \cite{viton-hd}, HR-VTON \cite{hr-vton} and  GP-VTON \cite{gp-vton}) and Diffusion-based VTON techniques (e.g., LaDI-VTON \cite{ladi-vton} and DCI-VTON \cite{dci-vton}) often fail to perfectly retain every appearance/texture detail of the given garment (e.g., the complex patterns or texts). Instead, our GarDiff exploits garment-focused diffusion process to preserve most of fine-grained details of the given garment, pursuing more controllable person image generation.}
    \label{Figure 1}

 \end{center}
To tackle VTON task, prior works \cite{flow1, 2018eccvtryon, han2018viton, viton-hd, cp-vton, dual-vton, 2018eccvtryon, fw-gan, vtnfp,2020tryoncvpr} commonly hinge on an explicit warping process to directly deform the appearance of in-shop garment conditioned on the pose of target person. However, this way is often prone to suffer from distortions and artifacts over warped garments, due to the misalignment between warped garment and  target person's body. To alleviate this misalignment issue, several subsequent works \cite{PF-AFN, FS-VTON, hr-vton, gp-vton, 2021cvprtryon} further upgrade warping process with an additional generative process. They capitalize on the typical Generative Adversarial Networks \cite{gan} to synthesize the final person image based on conditions like warped garment and human body. While effective, the synthesized person images still tend to be unsatisfactory in challenging cases (see the unrealistic artifacts in Figure~\ref{Figure 1} (a), (b) and (c)).
This might be attributed to the limited capacity of the underlying generative model (GAN) to synthesize person image with complex patterned garment and variable human pose.

Recently, Diffusion models \cite{ddpm1, latent-diff, ddpm2, song2019generative,qian2024boosting,zhu2024sd,zhang2024trip,chen2023control3d,chen2023controlstyle} emerge as a new trend of generative modeling in numerous image synthesis tasks, demonstrating better scalability and easier/stable training than GAN-based solutions. Motivated by this, recent advances \cite{dci-vton, ladi-vton} have been dedicated to remould pre-trained Latent Diffusion Model \cite{latent-diff} of text-to-image synthesis by leveraging warped garment as additional condition during diffusion process. Although promising results are attained, these diffusion-based VTON approaches (e.g., LaDI-VTON \cite{ladi-vton} and DCI-VTON \cite{dci-vton}) still fail to completely retain every detail of in-shop garment, especially for high-frequency texture details of complex patterns/texts (see Figure~\ref{Figure 1} (d) and (e)). We speculate that degenerated results might be caused by holistic diffusion process over latent space where compressed latent code is not capable of memorizing every intricate garment detail and providing sufficient guidance for VTON task.

In an effort to mitigate this problem, our work shapes a new way to upgrade the latent diffusion model with amplified guidance of both visual appearance and high-frequency texture details for VTON task. Technically, we propose a novel Garment-focused Diffusion model (GarDiff) to progressively excavate more prior knowledge about fine-grained garment details. Such prior knowledge acts as amplified garment-focused guidance to improve virtual try-on results. Specifically, CLIP \cite{clip} and Variational Auto-encoder (VAE) \cite{vae} are employed to encode the reference garment into appearance priors, which is regarded as additional conditions to guide diffusion process. To effectively leverage these priors, a new garment-focused adapter is introduced to the UNet of latent diffusion model. This design triggers the local fine-grained appearance alignment between the synthetic person image and reference garment \& human pose. Moreover, a novel appearance loss is defined on the warped reference garment to achieve the guidance of high-frequency prior, which is utilized to supervise the synthesis of high-frequency details in synthesized garment, pursuing better preservation of high-frequency texture details in VTON. Eventually, our GarDiff faithfully produces person images with better-aligned garment details (see Figure~\ref{Figure 1} (f)).

The main contribution of this work is the proposal of garment-focused diffusion model that facilitates virtual try-on tasks. This also leads to the elegant view of how a diffusion model should be designed for excavating the garment-focused prior knowledge (e.g., visual appearance and high-frequency texture details) tailored to VTON, and how to improve diffusion process with these amplified garment-focused guidance. Through an extensive set of experiments on VITON-HD and DressCode datasets, our GarDiff consistently achieves competitive performances against state-of-the-art VTON methods.

%% file: sec/2_relate_work.tex
\section{Related Work}
\label{sec:formatting}

\textbf{GAN-based Virtual Try-on.} To tackle VTON task, prior works \cite{flow1, 2018eccvtryon, han2018viton, viton-hd, cp-vton, dual-vton, 2018eccvtryon, fw-gan, vtnfp, PF-AFN, FS-VTON, hr-vton, gp-vton, 2021cvprtryon} capitalize on the GAN to synthesize the final person image based on the conditions like warped garment and human body. VITON~\cite{han2018viton} is a pioneering work that employs a refinement network to composite warped garments generated through TPS~\cite{tps} with the target person. CP-VTON~\cite{cp-vton} introduces an upgraded learnable TPS transformation for achieving more robust alignment between the target person and the garment. VITON-HD~\cite{viton-hd} designed an alignment-aware segment generator to fill the misaligned regions with the garment texture through multi-scale refinement. HR-VTON~\cite{hr-vton} proposes a novel try-on condition generator that unifies the warping and segmentation generation modules for handling the misalignment and occlusion. GP-VTON~\cite{gp-vton} presents an advanced LFGP warping module for creating deformed garments, which is optimized with a new DGT training strategy.

\noindent \textbf{Diffusion-based Virtual Try-on.} Recently, diffusion models \cite{ddpm1,ddpm2,gu2022vector} start to dominate in natural image generation due to its superior ability in generating high-fidelity realistic images compared to GAN-based models. Inspired by this, a series of diffusion-based virtual try-on models begin to emerge. TryOnDiffusion \cite{tryondiffusion} unifies two UNets to preserve garment details and warp the garment in a single network. LaDI-VTON \cite{ladi-vton} introduces the textual inversion component that maps visual features of reference garment to CLIP token embedding space as condition of diffusion model. DCI-VTON \cite{dci-vton} further use warping network to warp reference garment, which is fed into diffusion model as an additional guidance. Although promising results are attained, these diffusion-based VTON approaches still fail to completely retain every detail of the reference garment.





%% file: sec/3_method.tex
\section{METHOD}
\begin{figure*}[t]
    \centering
    \includegraphics[width=0.95\textwidth]{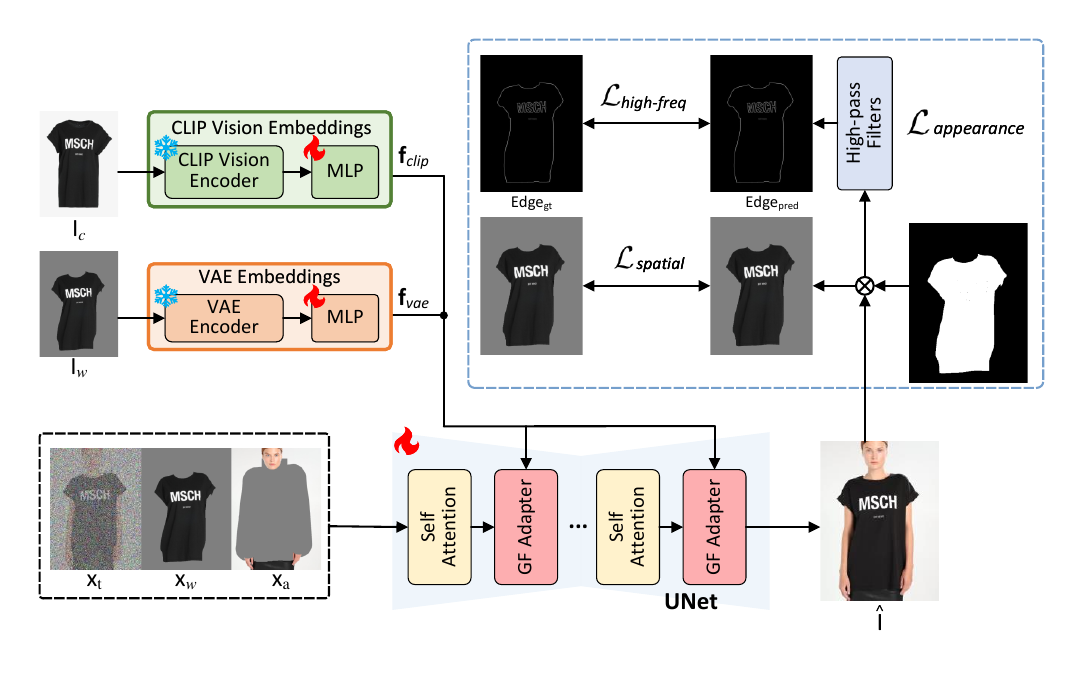}
    
    \caption{An overview of our GarDiff. The cross-attention layer is substituted with the garment-focused vision adapter in each Transformer block. First, we extract the CLIP visual embeddings $\mathbf{f}_{clip}$ and VAE embeddings $\mathbf{f}_{vae}$ of the target garment $\mathbf{I}_c$ and warped garment $\mathbf{I}_w$, respectively. Then the two embeddings are fed into the garment-focused adapter as keys and values via a decoupled cross-attention to guide the diffusion process for pursuing local fine-grained alignment with the appearance of target garment. Meanwhile, we employ a novel appearance loss $\mathcal{L}_{appearance}$ comprised of spatial perceptual loss $\mathcal{L}_{spatial}$ and high-frequency promoted loss $\mathcal{L}_{high\text{-}freq}$ over the generated garment to enhance the proficiency of GarDiff in generating high-frequency details. }
    \label{fig:framework}

\end{figure*}

\begin{figure}[t]
    \centering
    \includegraphics[width=0.98\textwidth]{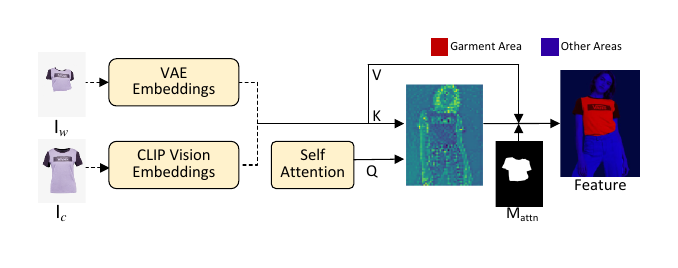}

    \caption{Implementation details of our garment-focused adapter. For the given target garment $\mathbf{I}_c$ and warped garment $\mathbf{I}_w$, the CLIP visual embeddings $\mathbf{f}_{clip}$ and VAE embeddings $\mathbf{f}_{vae}$ are extracted and fed into the garment-focused adapter as the keys and values through a decoupled cross-attention. $\mathbf{M}_{attn}$ is used to suppress the weights unrelated to garment area in the attention map for generating garment-focused features.}
    \label{mask}

\end{figure}

In this work, we propose a Garment-focused Diffusion model (GarDiff), an upgraded latent diffusion model with amplified guidance of both visual appearance and high-frequency texture details for VTON task. In this section, we will first provide a concise overview of our GarDiff, followed by the details of the proposed garment-focused vision adapter and appearance loss. 

\subsection{Overview}
\label{sec:tech_overview}
Generally, given a person image $\mathbf{I}_p\in{\mathbb{R}}^{H\times W\times3}$ and in-shop garment $\mathbf{I}_c\in{\mathbb{R}}^{H'\times W'\times3}$, our GarDiff is optimized to synthesize a high-quality realistic image $\mathbf{I}\in{\mathbb{R}}^{H\times W\times3}$, where the person wears the in-shop garment $\mathbf{I}_c$. To effectively leverage the appearance guidance of the given garment for high-fidelity person image generation, the original cross-attention layers in the UNet of diffusion model are substituted with our proposed garment-focused vision adapter modules. An overview of our GarDiff is illustrated in Figure \ref{fig:framework}.


In the forward diffusion process, similar to vanilla diffusion model, we gradually add noise to the target image $\mathbf{I}$ according to the Markov chain. Specifically, we first utilize the VAE encoder $\mathcal{E}(\cdot)$ to map the target image $\mathbf{I}$ to the latent space: $\mathbf{x}_0 = \mathcal{E}(\mathbf{I})$. Then we add noise $\epsilon$ to $\mathbf{x}_0$ at an arbitrary timestep $t \in [1, 1000]$ as follows:
\begin{equation}
\label{add_noise}
    \mathbf{x}_t = \sqrt{\alpha}_t\mathbf{x}_0 + \sqrt{1-\alpha_t}\epsilon,
\end{equation}
where ${\alpha}_t = \prod _{s = 1}^t(1 - {\beta _s})$, $\epsilon \sim \mathcal{N}(0,1)$, and $\beta _s$ is the pre-defined variance schedule at timestep $s$.

In the reverse diffusion process, we first warp the in-shop garment $\mathbf{I}_c$ 
 similar to \cite{gp-vton} to achieve the maximum conformity of the target garment with the body's posture, obtaining the warped garment $\mathbf{I}_w\in{\mathbb{R}}^{H'\times W'\times3}$ and warped mask $\mathbf{m}_w\in{\{0, 1\}}^{H\times W}$. The warped mask $\mathbf{m}_w$ indicates the area of the warped garment. To preserve the area unrelated to the garment, a garment-agnostic image $\mathbf{I}_a$ with the region intended for garment placement fully masked from the person image $\mathbf{I}_p$ is also extracted. Then, the noisy image latent $\mathbf{x}_t$, the latent warped garment map $\mathbf{x}_w = \mathcal{E}(\mathbf{I}_w)$ and the latent agnostic map $\mathbf{x}_a = \mathcal{E}(\mathbf{I}_a)$ are concatenated along the channel dimension, leading to the input of the UNet ${\epsilon}_{\theta}$ \cite{ronneberger2015unet} of the diffusion model:
\begin{equation}
\mathbf{\gamma} = Concat(\mathbf{x}_w, \mathbf{x}_a, \mathbf{x}_t).
\end{equation}
During denoising, the UNet is trained to predict the added noise $\epsilon$ conditioned on the target garment $\mathbf{I}_c$ and warped garment $\mathbf{I}_w$. The objective function is formed as the mean-squared loss:
\begin{equation}
\label{eq:mse}
    \mathcal{L}_{mse} = ||\epsilon - \epsilon_\theta(\mathbf{\gamma}, \mathbf{I}_c, \mathbf{I}_w, t)||_2^2.
\end{equation}

\subsection{Garment-Focused Adapter}
To retain the appearance details of the given garment, previous works \cite{dci-vton,ladi-vton,tryondiffusion} merely steer the diffusion process with the garment appearance captured by CLIP vision encoder. However, it is difficult for CLIP vision encoder to perceive the fine-grained details in the target garment, since CLIP is optimized for image-text alignment at a coarse level. In our experiments, we found that VAE, serving as the reconstruction module for stable diffusion, exhibits much stronger capabilities in preserving texture details in images than CLIP. Therefore, it would be beneficial to involve VAE as additional appearance prior into stable diffusion.

Technically, we replace the cross-attention layer with a vision adapter in each Transformer block. The vision adapter takes two appearance priors are inputs: 1) the VAE embeddings of the warpped rendition $\mathbf{I}_w$ of the target garment $\mathbf{I}_c$ for recovering the complex patterns in the synthesized image, 2) the CLIP visual embeddings of $\mathbf{I}_c$ for generating the holistic structure regardless of the imperfection of the warpped result $\mathbf{I}_w$. Specifically, given the target garment $\mathbf{I}_c$ and the warped garment $\mathbf{I}_w$, the CLIP visual embeddings $\mathbf{f}_{clip}$ and VAE embeddings $\mathbf{f}_{vae}$ are calculated as:
\begin{equation}
\begin{aligned}
    &{\mathbf{f}_{clip} = \mathbf{MLP}^{clip}(\mathbf{CLIP}_v(\mathbf{I}_c))}, \\
    &{\mathbf{f}_{vae} = \mathbf{MLP}^{vae}(\mathcal{E}(\mathbf{I}_w))},
\end{aligned}
\end{equation}
where $\mathbf{CLIP}_v$, $\mathbf{MLP}^{clip}$, $\mathbf{MLP}^{vae}$ are the CLIP vision encoder, multi-layer perceptrons for CLIP visual embeddings and VAE embeddings, respectively. Considering the two embeddings focus on different granularity levels of the target garment, a decoupled cross-attention mechanism is devised in the vision adapter to separate the cross-attention layers for the CLIP visual embeddings and VAE embeddings. Given the features $\mathbf{f}_q$ from the last self-attention layer, the vision adapter operates as follows:
\begin{equation}
\label{eq:adapter}
\centering
\begin{split}
\begin{aligned}
\mathbf{Z}_{attn} = Softmax(\frac{\mathbf{Q}(\mathbf{K}_{clip})^\top}{\sqrt{d}}) \mathbf{V}_{clip} &+ Softmax(\frac{\mathbf{Q}(\mathbf{K}_{vae})^\top}{\sqrt{d}}) \mathbf{V}_{vae}, \\
\mathbf{Q}=\mathbf{f}_q\mathbf{W}^q,~ \mathbf{K}_{clip} = \mathbf{f}_{clip}\mathbf{W}_{clip}^k,~ &\mathbf{V}_{clip} = \mathbf{f}_{clip}\mathbf{W}_{clip}^v, \\
\mathbf{K}_{vae} = \mathbf{f}_{vae}\mathbf{W}_{vae}^k,~ &\mathbf{V}_{vae} = \mathbf{f}_{vae}\mathbf{W}_{vae}^v,
\end{aligned}
\end{split}
\end{equation}
where $\mathbf{W}^q$, $\mathbf{W}_{*}^k$, $\mathbf{W}_{*}^v$ are the trainable projection matrices in the decoupled cross-attention of the vision adapter for queries, keys and values, respectively.


It is relatively easier for the model to achieve favorable results on the regions devoid of garments, such as the torso skin, since those regions usually exhibit simpler patterns or textures. Therefore, improving virtual try-on results hinges critically upon the restoration of the fine-grained details in the given garment. In light of this, we propose to amplify the garment-focused guidance in the diffusion process. Specifically, the vision adapter is further upgraded with a novel garment-focused attention, leading to a garment-focused (GF) adapter that aims to pursue local fine-grained alignment with the visual appearance of the reference garment and human pose. As illustrated in Figure \ref{mask}, the warped mask $\mathbf{m}_w$ is downsampled to an attention mask $\mathbf{M}_{attn}$ that matches the resolution of the corresponding attention layer in the GF adapter, which is leveraged to suppress the attention weights unrelated to garment area in the attention map. Hence, Equation (\ref{eq:adapter}) can be reformulated for the garment-focused adapter as follows:
\begin{equation}
\begin{split}
\begin{aligned}
\mathbf{Z}_{attn}^{mask}= &[softmax(\frac{\mathbf{Q}(\mathbf{K}_{clip})^\top}{\sqrt{d}})\odot \mathbf{M}_{attn}] \times \mathbf{V}_{clip} + \\ &[softmax(\frac{\mathbf{Q}(\mathbf{K}_{vae})^\top}{\sqrt{d}})\odot \mathbf{M}_{attn}] \times \mathbf{V}_{vae}.
\end{aligned}
\end{split}
\end{equation}

\subsection{Appearance Loss}
Generally, diffusion model is merely optimized with the mean-squared loss defined in Equation (\ref{eq:mse}), which treats all the regions of the synthesized image equally without emphasizing the texture details in the garment area, failing to generate the accurate garment patterns. As complicated details in the in-shop garment typically manifest as high-frequency components (i.e., edges), a novel appearance loss is proposed to enforce the synthesized garment to be geometrically consistent with the high-frequency details of the reference garment, achieving improved fidelity and fine-grained textures. The appearance loss, as a composite adaptation loss, can be decomposed into two components: a spatial perceptual loss $\mathcal{L}_{spatial}$ and a high-frequency promoted loss $\mathcal{L}_{high\text{-}freq}$. 

Specifically, we estimate the latent $\hat{\mathbf{x}}_0$ given the noisy one $\mathbf{x}_t$ and the predicted noise $\epsilon_t$ from the UNet at timestep $t$ by reversing the process in Equation (\ref{add_noise}):
\begin{equation}
    \hat{\mathbf{x}}_0 = \frac{\mathbf{x}_t - \sqrt{1-\alpha_t}\epsilon_t}{\sqrt{\alpha_t}}.
\end{equation}
The latent $\hat{\mathbf{x}}_0$ is further converted back to the pixel space by the VAE decoder $\mathcal{D}$, leading to the predicted image $\hat{\mathbf{I}}$.

\noindent\textbf{Spatial Perceptual Loss.} Inspired by the Deep Image Structure and Texture Similarity (DISTS) metric \cite{DISTS}, the spatial perceptual loss is designed to capture both the structural and textural disparities between the predicted and ground-truth images in a perceptual feature space beyond pixels:
\begin{equation}
    \mathcal{L}_{spatial} = \mathcal{L}_{DISTS}(\hat{\mathbf{I}}\odot \mathbf{m}_w,\mathbf{I}\odot \mathbf{m}_w).
\end{equation}
Note that we use the warped mask $\mathbf{m}_w$ to emphasize the garment area.

\noindent\textbf{High-Frequency Promoted Loss.}
To substantively enhance the model's proficiency in generating high-frequency details, we employ edge detection to extract high-frequency information. Technically, the horizontal and vertical Sobel kernels are adopted as the high-pass filters to extract the edge maps $\hat{\mathbf{I}}_h$ / $\mathbf{I}_h$ from the predicted/target image, respectively. Formally, the high-frequency promoted loss is defined as:
\begin{equation}
    \mathcal{L}_{high\text{-}freq} = ||\hat{\mathbf{I}}_h \odot \mathbf{m}_w - \mathbf{I}_h \odot \mathbf{m}_w ||^2.
\end{equation}
Finally, the UNet is optimized by the following improved objective function:
\begin{equation}
\label{eq:loss}
\begin{aligned}
    \mathcal{L} &= \mathcal{L}_{mse} + \lambda\mathcal{L}_{appearance}, \\
    \mathcal{L}_{appearance} &= \mathcal{L}_{spatial} + \mathcal{L}_{high\text{-}freq},
\end{aligned}
\end{equation}
where $\lambda$ is the hyper-parameter used to balance the mean-squared loss and the proposed appearance loss.


%% file: sec/4_exp.tex
\section{Experiments}

\subsection{Experimental Settings}
\noindent\textbf{Dadasets.} We empirically verify and analyze the effectiveness of our GarDiff on two popular virtual try-on datasets, VITON-HD \cite{viton-hd} and DressCode \cite{dresscode}. The VITON-HD dataset \cite{viton-hd} comprises 13,679 frontal-view woman and upper garment image pairs. In line with the general practices of the previous works \cite{dci-vton, ladi-vton}, the dataset is divided into two disjoint subsets: a training set with 11,647 pairs and a test set with 2,032 pairs. The DressCode dataset consists of 53,795 image pairs, which are categorized into three macro-categories: 15,366 for upper-body clothes, 8,951 pairs lower-body clothes and 29,478 for dresses. Following the original splits, 1,800 image pairs from each category are reserved for test while the remaining image pairs are utilized for training. The experiments on DressCode and VITON-HD are conducted at the resolution of 512 $\times$ 384.

\noindent\textbf{Evaluation Metrics.}
We evaluate our GarDiff in both paired and unpaired settings following the virtual try-on literature. In the paired setting, the input garment corresponds to the one originally depicted in the person image. Following the standard evaluation setup, Structural Similarity (SSIM)\cite{wang2004image} and Learned Perceptual Image Patch Similarity (LPIPS)\cite{zhang2018unreasonable} are adopted to measure the similarity between the generated image and the ground-truth one. Additionally, the Fr\'echet Inception Distance(FID)\cite{fid} and Kernel Inception Distance(KID)\cite{kid} are employed to measure the quality and realism of the generated images. In the unpaired setting, where the garment of the person image is changed to a different one and the ground truth is unavailable, we report the performances of GarDiff in terms of FID and KID.

\noindent\textbf{Implementation Details.}
Our GarDiff is initialized from the pre-trained Stable Diffusion 2.1 and finetuned on the virtual try-on datasets. AdamW \cite{adam} ($\beta_1=0.9$, $\beta_2=0.999$) is employed to optimize the model for 200k steps. The learning rate is set to 0.00005 with linear warmup of 500 iterations. The hyper-parameter $\lambda$ in Equation (\ref{eq:loss}) and the weight decay are set to 0.001 and 0.01, respectively. OpenCLIP ViT-H/14 \cite{openclip} is utilized to extract the CLIP visual embeddings of the target garment. To enable classifier-free guidance \cite{classifier}, the embeddings of the garment are randomly dropped with a probability of 0.05. We train GarDiff on 4 NVidia RTX4090 GPUs for about four days and the model size is 5.15 GB. During inference, the image is progressively generated over 100 steps with a DDIM \cite{ddim} sampler, and the scale of classifier-free guidance is set to 7.5 by default.


\subsection{Quantitative Results}

\begin{table}[t] \scriptsize
    \centering
    \setlength\tabcolsep{1.0pt}
    \caption{Quantitative performance comparisons on VITON-HD dataset. $\mathbf{FID_p}$/$\mathbf{KID_p}$ stands for the $\mathbf{FID}$/$\mathbf{KID}$ score in paired setting, while $\mathbf{FID_u}$/$\mathbf{KID_u}$ stands for the $\mathbf{FID}$/$\mathbf{KID}$ score in unpaired setting. Note that the KID score is multiplied by 100.}
    \begin{tabular}{cccccccc}
    \toprule
   \textbf{Model} & $\mathbf{SSIM\uparrow}$ &  $\mathbf{LPIPS\downarrow}$ &  $\mathbf{FID_p\downarrow}$ &  $\mathbf{KID_p\downarrow}$ &$\mathbf{FID_u\downarrow}$ &  $\mathbf{KID_u\downarrow}$ \\
    \midrule
   VITON-HD~\cite{viton-hd}& 0.843&  0.076&  -&  -&  11.64& 0.300\\
   PF-AFN~\cite{PF-AFN}& 0.858&  0.082&  -&  -&  11.30& 0.283\\
   HR-VTON~\cite{hr-vton}&  0.878&  0.061&  -&  -&  9.90& 0.188\\
   GP-VTON~\cite{gp-vton}&  0.894&  0.080&  -&  -&  9.20& -\\
    \midrule
   Paint-by-Example~\cite{paint-by-example}&  0.843&  0.087&  -&  -&  10.15&0.204\\
   LaDI-VTON~\cite{ladi-vton}&  0.879&  0.059&  6.66&  0.108&  9.41& 0.167\\
   DCI-VTON~\cite{dci-vton}&  0.896& 0.043&  -&  -&  8.09& 0.028\\
   \textbf{GarDiff}&  \textbf{0.912}& \textbf{0.036}&  \textbf{6.02}&  \textbf{0.019}&\textbf{7.89}&\textbf{0.027} \\
    \bottomrule
    \end{tabular}
    \label{table-viton}
\end{table}

\begin{table}[t] \scriptsize
    \centering
    \setlength\tabcolsep{1.0pt}
    \caption{Quantitative performance comparisons on DressCode dataset. $\mathbf{FID_p}$/$\mathbf{KID_p}$ stands for the $\mathbf{FID}$/$\mathbf{KID}$ score in paired setting, while $\mathbf{FID_u}$/$\mathbf{KID_u}$ stands for the $\mathbf{FID}$/$\mathbf{KID}$ score in unpaired setting. Note that the KID score is multiplied by 100.}
    \begin{tabular}{cccccccc}
    \toprule
    \textbf{Dataset} & \multicolumn{6}{c}{\textbf{DC\_upper-body}}\\ \cmidrule(lr){1-1} \cmidrule(lr){2-7}  
   \textbf{Model} & $\mathbf{SSIM\uparrow}$ &  $\mathbf{LPIPS\downarrow}$ &  $\mathbf{FID_p\downarrow}$ &  $\mathbf{KID_p\downarrow}$ &$\mathbf{FID_u\downarrow}$ &  $\mathbf{KID_u\downarrow}$ \\
    \midrule
  CP-VTON+~\cite{cp-vton} &0.918&0.078&19.70&1.16&22.18&12.09\\
   PF-AFN~\cite{PF-AFN} &0.918&-&-&-&14.32&-\\
   PSAD~\cite{dresscode}&0.938&0.049&13.87&0.640&17.51&7.15\\
   GP-VTON~\cite{gp-vton}&0.947&0.036&-&-&11.98&-\\
    \midrule
   LaDI-VTON~\cite{ladi-vton}&0.928&0.049&9.53&0.198&13.26&2.67\\
   DCI-VTON~\cite{dci-vton}&-& \textbf{0.030}&-&-&\textbf{10.82}&-\\
   \textbf{GarDiff}&  \textbf{0.952}& \textbf{0.030}&  \textbf{8.69}&  \textbf{0.144}&11.32&\textbf{2.34} \\
    \bottomrule
    \end{tabular}

    \begin{tabular}{cccccccc}
    \toprule
    \textbf{Dataset} & \multicolumn{6}{c}{\textbf{DC\_lower-body}} \\ \cmidrule(lr){1-1} \cmidrule(lr){2-7}  
   \textbf{Model} & $\mathbf{SSIM\uparrow}$ &  $\mathbf{LPIPS\downarrow}$ &  $\mathbf{FID_p\downarrow}$ &  $\mathbf{KID_p\downarrow}$ &$\mathbf{FID_u\downarrow}$ &  $\mathbf{KID_u\downarrow}$ \\
    \midrule
   CP-VTON+~\cite{cp-vton} &0.913&0.083&-&-&18.85&10.24\\
   PF-AFN~\cite{PF-AFN} &0.907&-&-&-&18.32&-\\
   PSAD~\cite{dresscode}&0.932&0.051&13.14&0.559&19.68&8.90\\
   GP-VTON~\cite{gp-vton}&\textbf{0.941}&0.042&-&-&16.07&-\\
    \midrule
   LaDI-VTON~\cite{ladi-vton}&0.922&0.051&8.52&0.104&14.80&3.13\\
   DCI-VTON~\cite{dci-vton}&-& \textbf{0.035}&-&-&12.34&-\\
   \textbf{GarDiff}&  0.939& \textbf{0.035}&  \textbf{8.01}&  \textbf{0.090}&\textbf{12.29}&\textbf{2.88} \\
    \bottomrule
    \end{tabular}

 \begin{tabular}{cccccccc}
    \toprule
    \textbf{Dataset} & \multicolumn{6}{c}{\textbf{DC\_dresses}} \\ \cmidrule(lr){1-1} \cmidrule(lr){2-7}  
   \textbf{Model} & $\mathbf{SSIM\uparrow}$ &  $\mathbf{LPIPS\downarrow}$ &  $\mathbf{FID_p\downarrow}$ &  $\mathbf{KID_p\downarrow}$ &$\mathbf{FID_u\downarrow}$ &  $\mathbf{KID_u\downarrow}$ \\
    \midrule
   CP-VTON+~\cite{cp-vton} &0.863&0.123&18.75&1.11&21.83&12.31\\
   PF-AFN~\cite{PF-AFN} &0.869&-&-&-&13.59&-\\
   PSAD~\cite{dresscode}&0.885&0.074&12.38&0.468&17.07&6.66\\
   GP-VTON~\cite{gp-vton}&0.886&0.072&-&-&12.26&-\\
    \midrule
   LaDI-VTON~\cite{ladi-vton}&0.868&0.089&9.07&0.112&13.40&2.50\\
   DCI-VTON~\cite{dci-vton}&-& 0.068&-&-&12.25&-\\
   \textbf{GarDiff}&  \textbf{0.891}& \textbf{0.065}&  \textbf{8.77}&  \textbf{0.096}&\textbf{12.05}&\textbf{2.09} \\
    \bottomrule
    \end{tabular}

    \label{table-DC}
\end{table}

\noindent\textbf{VITON-HD.}
We compare our GarDiff with a series of state-of-the-art virtual try-on methods including GAN-based methods (VITON-HD \cite{viton-hd}, PF-AFN \cite{PF-AFN}, HR-VTON \cite{hr-vton}, GP-VTON \cite{gp-vton}) and Diffusion-based methods (DCI-VTON \cite{dci-vton}, Paint-by-Example \cite{paint-by-example} and LaDI-VTON \cite{ladi-vton}). Table \ref{table-viton} summarizes the performance comparisons on the VITON-HD dataset. It can be easily observed that our proposed GarDiff consistently demonstrates superior performances compared to the other methods across all the evaluation metrics. In particular, GP-VITON improves VITON-HD by introducing the try-on condition generator that serves as a unified module in warping and segmentation generation stages. Paint-by-Example further boosts the performances by framing the VTON task as exemplar-based image inpainting and filling the target region of source image with the garment in the reference image. By exploiting the textual inversion to maintain the details of the in-shop garment, LaDI-VTON exhibits better performance than Paint-by-Example. Moreover, DCI-VTON leverages a warping network to guide the image generation of diffusion model, leading to clear performance boosts. However, these existing approaches merely steer the generation process at a coarse level, and thus fail to perfectly retain the details of the given garment. On the contrary, our GarDiff facilitates the preservation of the garment's appearance by excavating the garment-focused prior knowledge and strengthening diffusion process with these amplified garment-focused guidance, achieving much better results in VTON task. Specifically, our GarDiff achieves 0.912 on SSIM score and makes a relative improvement of 1.78\% against the best competitor DCI-VTON.

\noindent\textbf{DressCode.}
Table \ref{table-DC} summarizes the performance comparisons on the DressCode dataset. Similar to the observations on VITON-HD, our proposed GarDiff surpasses the performances of the other competing methods across all the three macro-categories, which again evinces the pivotal merit of the garment-focused appearance guidance for preserving fine-grained garment attributes in the generated images. Particularly, our GarDiff leads to the relative improvements over LaDI-VTON by 2.58\% on SSIM for upper-body settings.

\begin{figure}[t]
    \centering
    \includegraphics[width=\textwidth]{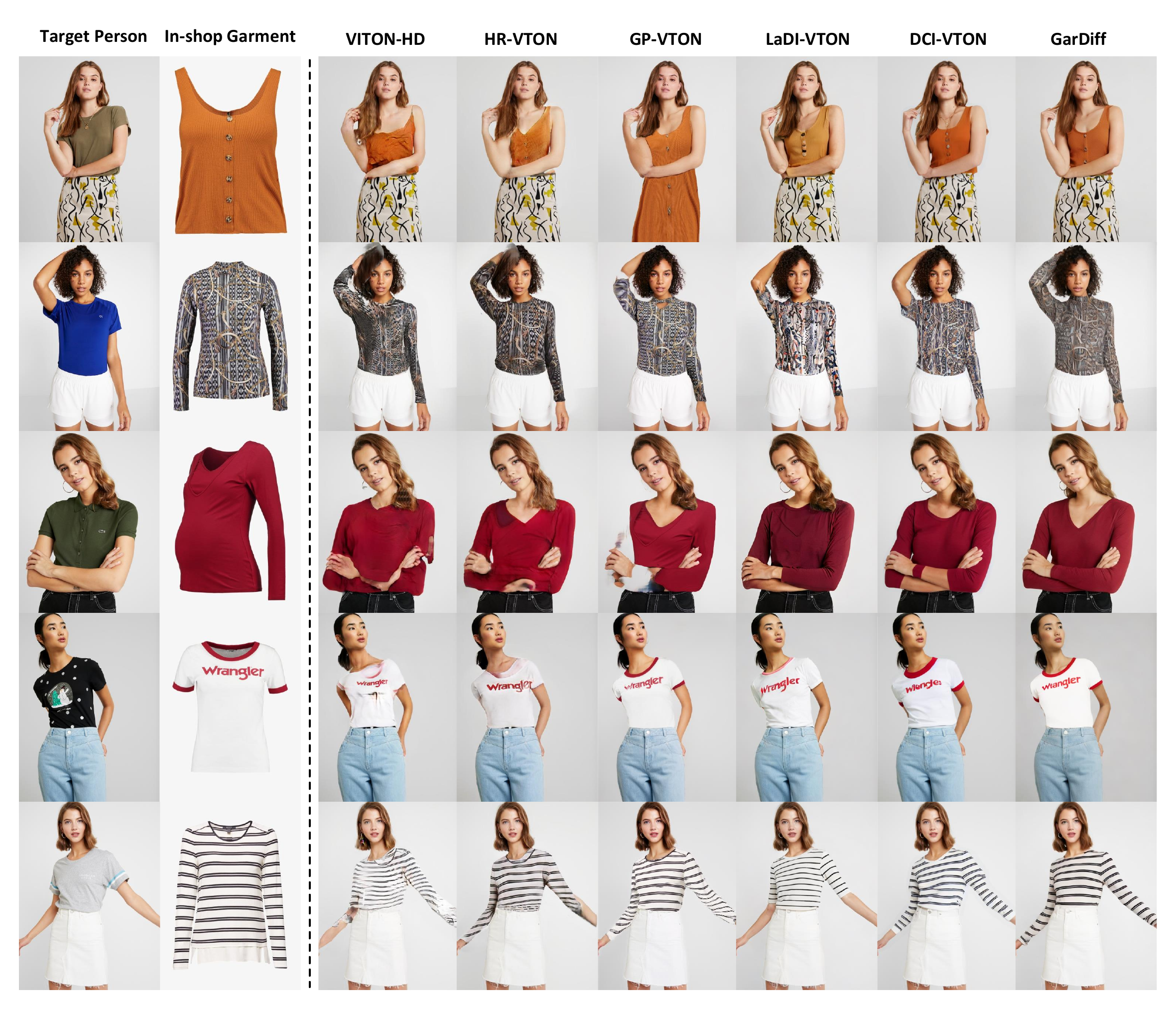}
    \caption{Examples generated by VITON-HD, HR-VTON, GP-VTON, LaDI-VTON, DCI-VTON and our GarDiff.}
    \label{qualitative}
\end{figure}

\begin{figure}[t]
    \centering
    \includegraphics[width=0.8\textwidth]{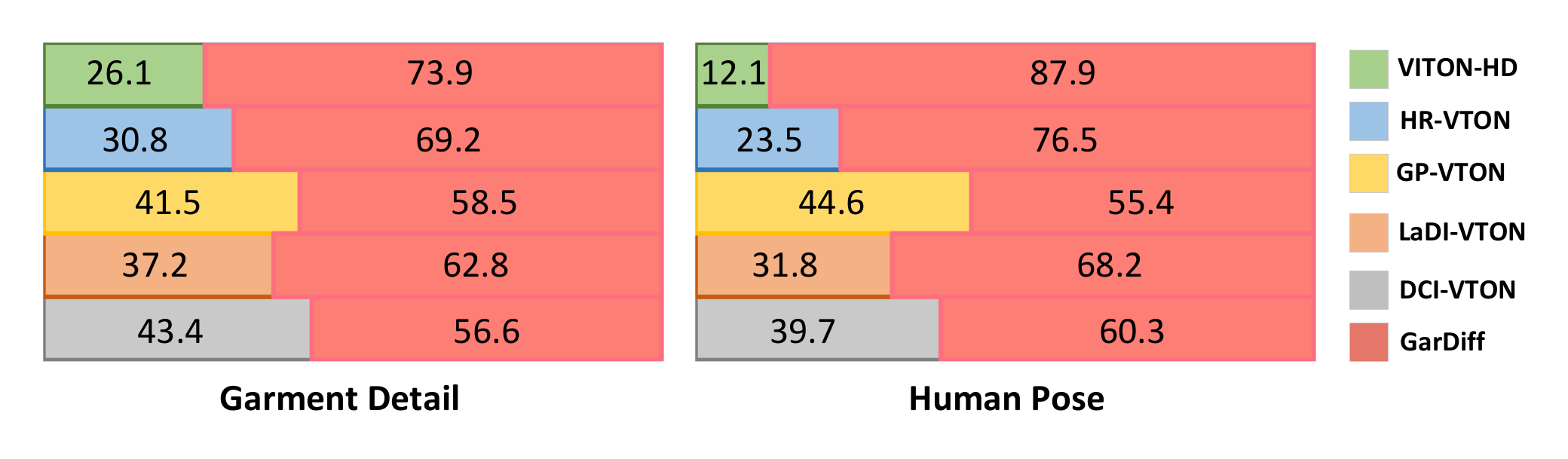}

    \caption{User study on 100 garment-person pairs randomly sampled from VITON-HD.}
    \label{exp-user_study}
  \end{figure}

\begin{table}[t] \scriptsize
    \centering
    \setlength\tabcolsep{1.0pt}
    \caption{Ablation study of our proposed GarDiff on VITON-HD dataset. \textbf{Base}: based model; \textbf{GFA}: garment-focused adapter; \textbf{AL}: appearance loss.}
    \begin{tabular}{lccccccc}
    \toprule
   \textbf{Model} & $\mathbf{SSIM\uparrow}$ &  $\mathbf{LPIPS\downarrow}$ &  $\mathbf{FID_p\downarrow}$ &  $\mathbf{KID_p\downarrow}$ &$\mathbf{FID_u\downarrow}$ &  $\mathbf{KID_u\downarrow}$ \\
    \midrule
   \textbf{Base}& 0.809&  0.117&  10.03&  0.190&  10.96& 0.473\\
   \textbf{Base+AL}& 0.829&  0.090&  8.72&  0.103&  10.22& 0.300\\
   \textbf{Base+GFA}&  0.898&  0.052&  6.61&  0.047&  8.21&             
         0.069\\
   \textbf{Base+GFA+AL}&  \textbf{0.912}& \textbf{0.036}&  \textbf{6.02}&  \textbf{0.019}&\textbf{7.89}&\textbf{0.027} \\
    \bottomrule
    \end{tabular}
    \label{tab_ablation}
\end{table}

\subsection{Qualitative Results}
Figure \ref{qualitative} showcases several virtual try-on results of different methods, coupled with the input person image and in-shop garment. As evidenced by the presented exemplar results, all the approaches demonstrate a certain degree of proficiency in resembling the appearance and texture details of the input in-shop garments. Specifically, Diffusion-based methods (LaDI-VTON, DCI-VTON and our GarDiff) consistently outperform the GAN-based approaches (VITON-HD, HR-VTON and GP-VTON) by leveraging the strong capability of diffusion models in generating high-fidelity images. Despite both DCI-VTON and our GarDiff capitalizing on diffusion models, the former approach achieves inferior results to the latter. The underlying rationale lies in the fact that DCI-VTON guides the diffusion process with CLIP visual embeddings of the garment only, while our GarDiff additionally leverages the appearance prior from the VAE encoder through a vision adapter to better preserve the fine-grained details. Moreover, GarDiff is further upgraded with garment-focused attention machanism and optimized with a new appearance loss to steer the diffusion process with amplified garment-focused guidance, yielding images with enhanced local alignment to the appearance of the garment. For example, our GarDiff better restores the text ``wrangler'' on the garment than DCI-VTON in the fourth row.


Furthermore, we conduct human study to compare our GarDiff against five strong baselines (HR-VTON, VITON-HD, GP-VTON, Ladi-VTON, DCI-VTON) over 100 randomly sampled garment-person pairs in VITON-HD (unpaired setting). 10 evaluators from diverse education background are invited to rank the best VTON result between our GarDiff and the five competing methods based on two criteria: (1) garment detail preservation, (2) human pose alignment. Figure \ref{exp-user_study} shows the percentages of top-1 ranking for each method, and our GarDiff achieves the best results in both evaluated dimensions.


\subsection{Analysis and Discussions}
\subsubsection{Ablation Study on GarDiff.} We conduct an ablation study to investigate how each design in our GarDiff influences the overall performances on the VITON-HD dataset. Table \ref{tab_ablation} details the performance comparisons among different ablated runs of our GarDiff. We start from the stable diffusion inpainting as our base model (\textbf{Base}), which replaces the original CLIP text encoder with the CLIP vision encoder. \textbf{Base+GFA} further boosts \textbf{Base}) by concurrently leveraging the CLIP visual embeddings and VAE embeddings to guide the diffusion process, which verifies the merit of applying the appearance prior from VAE. It is not surprising that \textbf{Base+AL} outperforms \textbf{Base} by supervising the training with the novel appearance loss that facilitates the preservation of garment detail. Finally, our full GarDiff (\textbf{Base+GFA+AL}) achieves the best performances through the synergetic integration of the two proposed components.



\begin{figure}[t]
    \centering
    \includegraphics[width=\textwidth]{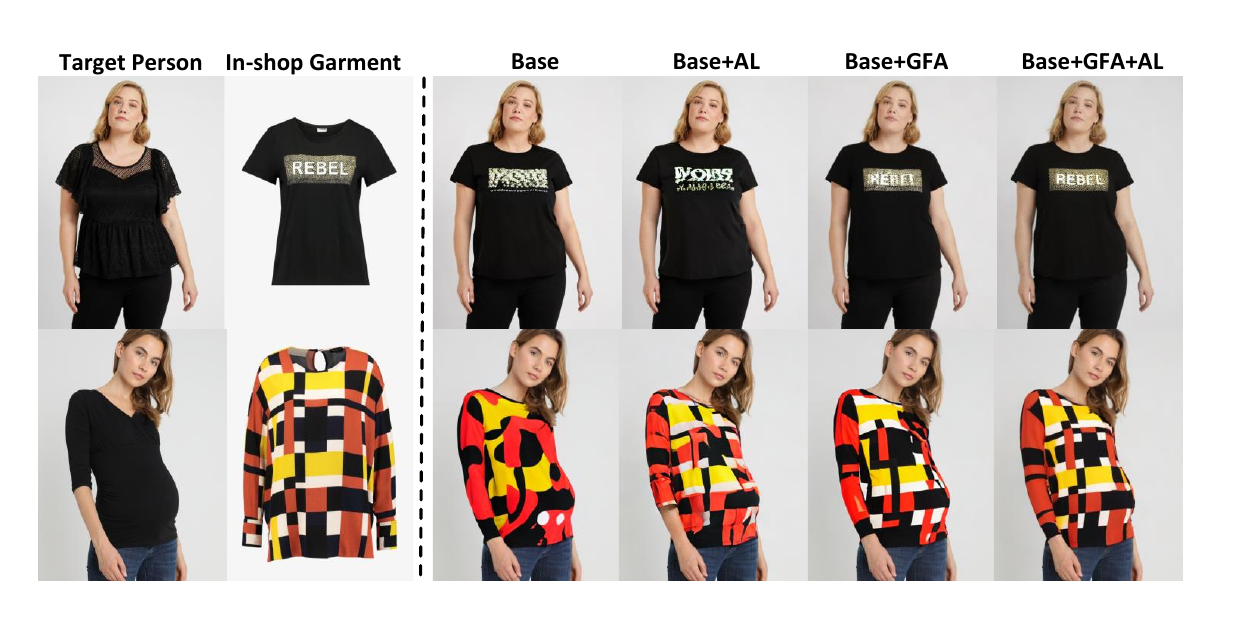}

    \caption{Ablation study on the pivotal components of GarDiff.}
    \label{exp-ablation}
  \end{figure}

To illustrate the impact of these components more intuitively, we visualize these ablated runs in Figure \ref{exp-ablation}. Compared with other ablated runs, our final model \textbf{Base+GFA+AL} (i.e., GarDiff) can preserve most of the fine-grained details of the given garments. Take the first case as an example, compared to the base model \textbf{Base}, \textbf{Base+GFA} effectively retains texture details of the letters. Similarly, when the garment-focused vision adapter is additionally integrated into the \textbf{Base} model, the generated results are further improved by \textbf{Base+AL}. 

\begin{figure}[t]
    \centering
    \includegraphics[width=\textwidth]{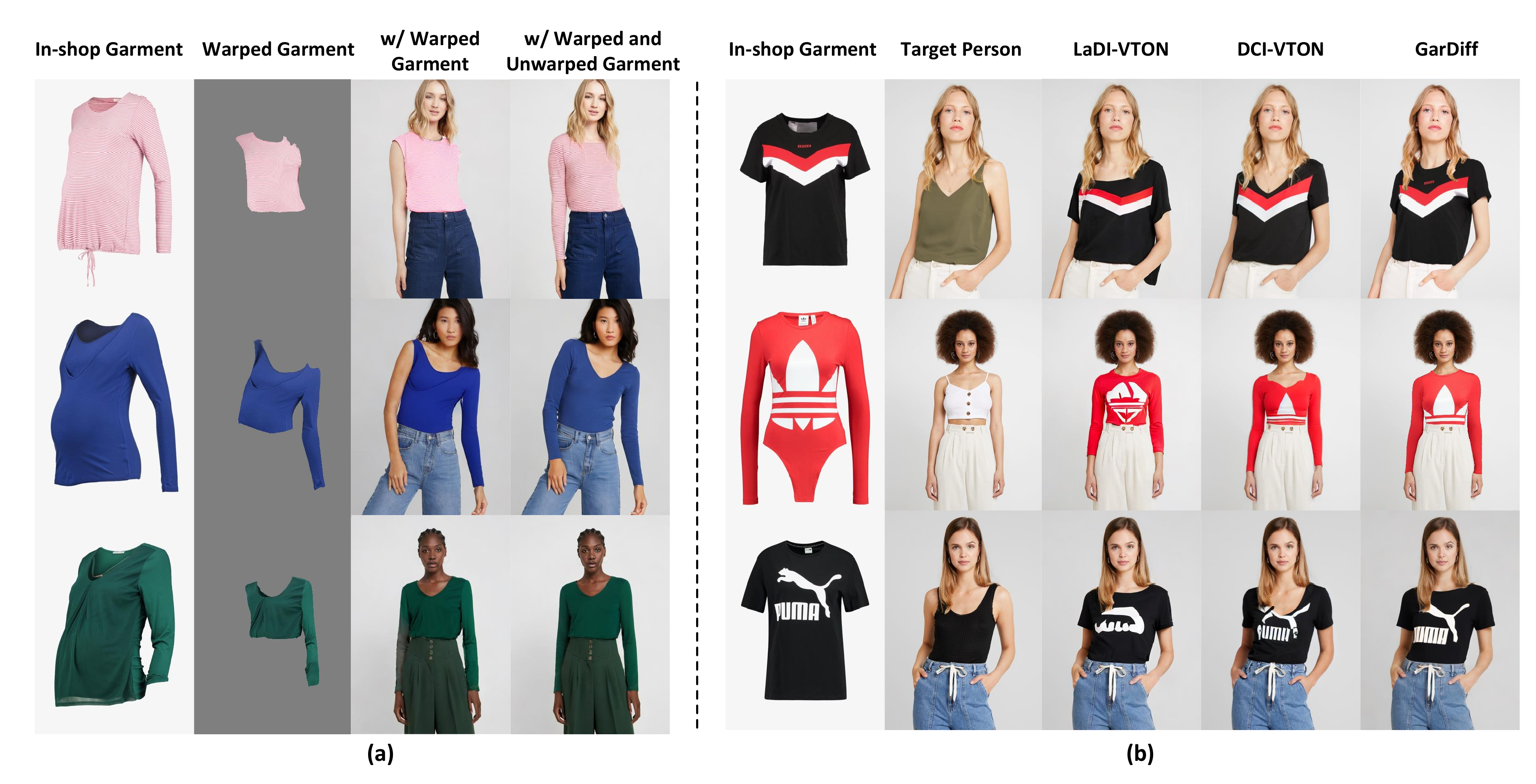}

    \caption{(a) Examples generated by our GarDiff with or without unwarpped garment. (b) Comparisons between Diffusion-based baselines (LaDI-VTON and DCI-VTON) and our GarDiff regarding the preservation of fine-grained details.}

    \label{exp-ablation-warp}
  \end{figure}

 \subsubsection{Effect of Unwarpped Garment.}
In contrast to the Diffusion-based DCI-VTON that merely captializes on the warpped garment derived from an warping networks for VTON, our GarDiff additionally incorporates the unwarpped garment to faithfully retain the appearance and shape details of the reference garment. In this way, high-quality VTON results can be achieved even when the warpped garments predicted from the warping networks are defective. Some exampels are shown in Figure \ref{exp-ablation-warp}(a). Regarding the long-sleeve T-shirt on th first row, the warpped garment fails to accurately restore the shape of the input garment with one of the sleeves missing. As a result, less favorable outcomes are obtained by solely leveraging the erroneously warped garment. Our GarDiff, employing both the appearance priors over the warpped and unwarpped garments, generates high-fidelity images.

\subsubsection{Preservation of Fine-grained Details.}
With the assistance of the proposed garment-focused attention mechanism and appearance loss, our GarDiff is capable of accurately aligning the visual appearance of the garment in the generated samples with the reference one. Figure \ref{exp-ablation-warp}(b) showcases the images synthesized by Diffusion-based competing methods (LaDI-VTON and DCI-VTON) and our proposed GarDiff, involving the cases with fine-grained details. Compared to LaDI-VTON and DCI-VTON which fail to achieve satisfactory results, our GarDiff successfully restores the small letters in the garment on the first row.

%% file: sec/5_conclusion.tex
\section{Conclusion}
In this work, we have presented the Garment-focused Diffusion model (GarDiff) that is capable of preserving the fine-grained details of the target garment in the virtual try-on task. Specifically, GarDiff remoulds the pre-trained latent diffusion model with appearance priors from the CLIP vision encoder and the VAE encoder for the reference garment and then integrates these priors into UNet through a garment-focused vision adapter. In this way, the diffusion process is effectively strengthened with the amplified appearance guidance from the given garment. A novel appearance loss is further devised to enforce the synthesized garment to be consistent with the high-frequency details and the geometric shape of target garment. Extensive experiments conducted on VITON-HD and DressCode datasets demonstrate the superiority of our GarDiff. More remarkably, we achieve new state-of-the-art performances on the two virtual try-on datasets.


\subsubsection{Broader Impact.}
Recent advances in generative modeling offer new possibilities for creating and manipulating digital media but also pose risks of generating deceptive content. Our proposed GarDiff might be nefariously used to ``undress'' individuals by substituting their original attire with undergarments for pornographic applications, and we emphatically denounce any such activities.
